\documentclass[journal]{IEEEtran}
\usepackage{cite}

\ifCLASSINFOpdf
   \usepackage[pdftex]{graphicx}
   \DeclareGraphicsExtensions{.pdf,.jpeg,.png,.jpg}
\else
   \usepackage[dvips]{graphicx}
\fi
\usepackage[cmex10]{amsmath}
\usepackage[tight,footnotesize]{subfigure}

\usepackage{stfloats}
\usepackage{url}

\usepackage{amssymb}
\usepackage{amsthm}
\newcommand*\norm[1]{\left\| #1 \right\|}

\begin{document}
%
\title{Multiple Closed-Form Local Metric Learning for K-Nearest Neighbor Classifier}
%
%
%

\author{Jianbo Ye \thanks{Jianbo Ye is with the Department of Computer Science, The University of Hong Kong, e-mail: \texttt{jbye@cs.hku.hk}.}
}

\maketitle

\begin{abstract}
Many researches have been devoted to learn a Mahalanobis distance metric, which can effectively improve the performance of kNN classification. Most approaches are iterative and computational expensive and linear rigidity still critically limits metric learning algorithm to perform better. We proposed a computational economical framework to learn multiple metrics in closed-form.
\end{abstract}


%
\IEEEpeerreviewmaketitle

\section{Introduction}
%
%
%
%
\IEEEPARstart{K}{-nearest} neighbor algorithm(kNN)\cite{cove67} is the oldest and simplest methods for classifying objects based on closest training examples in the feature space. Despite its simplicity. the kNN rule often yields competitive results. As a type of instance-based learning, kNN rules in effect compute the decision boundary in an implicit manner.

Most implementation of kNN compute simple Euclidean distances(assuming the examples are represented as vector inputs), when no prior knowledge is available. Unfortunately, Euclidean distance ignores any statistical regularities that might be estimated from a large training set of labeled examples. Motivated by these issues, a number of researchers (such as Xing\cite{XingNJR02}, LDA\cite{Fisher1962}\cite{SCHHoi06}, RCA\cite{Bar-HillelHSW03}, NCA\cite{GoldbergerRHS04}, LMNN\cite{WeinbergerBS05}\cite{WeinbergerS08}, MCML\cite{GlobersonR06}, ITML\cite{DavisKJSD07}, BoostMetric\cite{ShenKWHJ09}) have demonstrated that kNN classification can be greatly improved by learning a Mahalanobis distance(quadratic metric) over the input space, which later was formally called \textit{distance metric learning} in literature. Most approaches in metric learning convert the learning process to a semi-definite programming(SDP). Some are solved by iterative numerical solvers, while others (were later proven to) like Xing, RCA and LDA, have closed-form solutions\cite{Alipanahi08}.

There are two major concerns in previous approaches of metric learning. Firstly, iterative solver computationally converge a metric learning process to a local optimal, such as NCA\cite{GoldbergerRHS04} and its derivatives like LDM\cite{Yang06}. Meanwhile even problem are formulated into a convex optimization and solved by SDP like MCML, ITML and LMNN, there exists a target draft between the objective function and kNN classifier's accuracy. Secondly, whatever formulation of objective in metric learning, the overall performance is restricted to linearity of Mahalanobis distance metric inherently. Hence for data sets which have intrinsic non-linear structures, nonlinear approaches like SVM\cite{Cortes95} or kernel learning are expected to perform better at a computational expense.

In this paper, we proposed a novel computationally economical framework to learning multiple metrics, which benefits from the closed-form method like Linear Discriminant Analysis(LDA\cite{Fisher1962}) in their efficiency, explores the margin discriminative power like Large Margin Nearest Neighbor(LMNN\cite{WeinbergerBS05}), and leverage the restriction of linearity and instance-based query efficiency by learning multiple metrics.

Our paper is organized as follows: section~\ref{frame} introduces the overall framework of our approach, named \textit{multiple closed-form local metric learning}(CFLML); section~\ref{detail} runs over all necessary technical details in CFLML; section~\ref{impl} gives a brief note on efficient implementation; section~\ref{result} provides experimental results of CFLML and some other approaches in literature; we concludes our paper with a discussion on future works in section~\ref{concls}.


\section{Framework overview}\label{frame}
\subsection{K-nearest neighbor classification}
In the classification phase, $k$ is a user-defined constant, which is chosen heuristically to achieve optimal. The best $k$ basically is determined by the statistical properties of labeled instances, where large values of $k$ reduce the effect of noise on the classification, but make boundaries between classes less distinct. Hence it is somewhat wise to select a larger $k$ if different classes in training set are widely separated. As a comment, some techniques like LMNN are implicitly designed to be applicable for small $k$($k=1,3$).
\subsection{Mahalanobis metric}
In short, the Mahalanobis distance of multivariate column vectors $\mathbf{x}, \mathbf{y}\in \mathbb{R}^n$ and a covariance matrix $A_{n\times n}$(positive semi-definite) is defined as
\[d(\mathbf{x},\mathbf{y}) = \sqrt{(\mathbf{x}-\mathbf{y})^TS(\mathbf{x}-\mathbf{y})},\]
where square matrix $A$ is guaranteed to be positive semi-definite, hence it has eigendecomposition
\[A = U^T\Lambda U = L^TL,\]
where $\Lambda_{m\times m}$ is a diagonal matrix formed by non-zero eigenvalues of $A$, and the columns of $U$ are the corresponding  eigenvectors. Set $L_{m\times n} = \Lambda^{1/2}U$, as we see the Mahalanobis distance metric is equivalent to apply a linear transform $L$ over the original vector space.
\subsection{Multiple local metrics}
A major limitation of Mahalanobis metric learning is that, it would preserve linear rigidity of the data set. Previous approaches are mainly supposed to make a trade-off when forming the problem to optimize a certain objective function, while enlarge the distance between different labeled pairs and shrink or preserve the distance between pairs of same labels. However, due to the linear rigidity, the trade-off is crucial to the performance of derived kNN classifier.

Note that as an extension to LMNN\cite{WeinbergerBS05}, a multiple metric approaches(MM-LMNN) is proposed in the same paper. Its general idea is to divide the train set into multiple clusters and learn multiple LMNNs individually. However in most cases, this approach does not significantly improve the overall performance.

In stead of dividing train set into clusters, which can be regarded a complementary separation in feature domain, we define scalar valued functions (in terms of metrics) to describe labelling ambiguity over feature domain, and associate instances with different metrics by selecting one with least ambiguity.
\subsection{Overview}
The key idea in our framework is that by providing a group of metrics(includes at least one metric) for instances, which we called parents, we could produced a child metric in complementary to the performance of parents. The offspring procedure is a closed-form solution of metric learning process, which is effective to improve the performance of kNN with combination of its parents and much computationally cheaper comparing to other iterative solver based methods.

In this paper, we use a simple stochastic local search as follows:
\begin{enumerate}
\item Set train set and validation set, and an initial metric $L_0$, target group of metrics $G=\{L_0\}$. Start iterations.
\item In $i$-th step, produce child $L_{i+1}$ from parent $\{L_i\}$.
\item If $\{L_{i+1}\}\cup G$ perform better than $G$, add $L_{i+1}$ into $G$, set backtrace\_count:=0; else backtrace\_count++;
\item The iteration stops when backtrace\_count reaches its maximum, output $G$.
\end{enumerate}

The above algorithm is supposed to be a \textit{radical} strategy. As a \textit{conservative} alternative, in second step we could produce the child $L_{i+1}$ from parents $\{L_i\}\cup G$. Note that in fact, we could formulate our problem as a standard evolutionary computation, while preserving the group size of $G$.
\section{Learning boundary-based discriminant kNN classifier}\label{detail}
\begin{figure}[htp]
\centering
\subfigure[Null]{\includegraphics[height=1.2in]{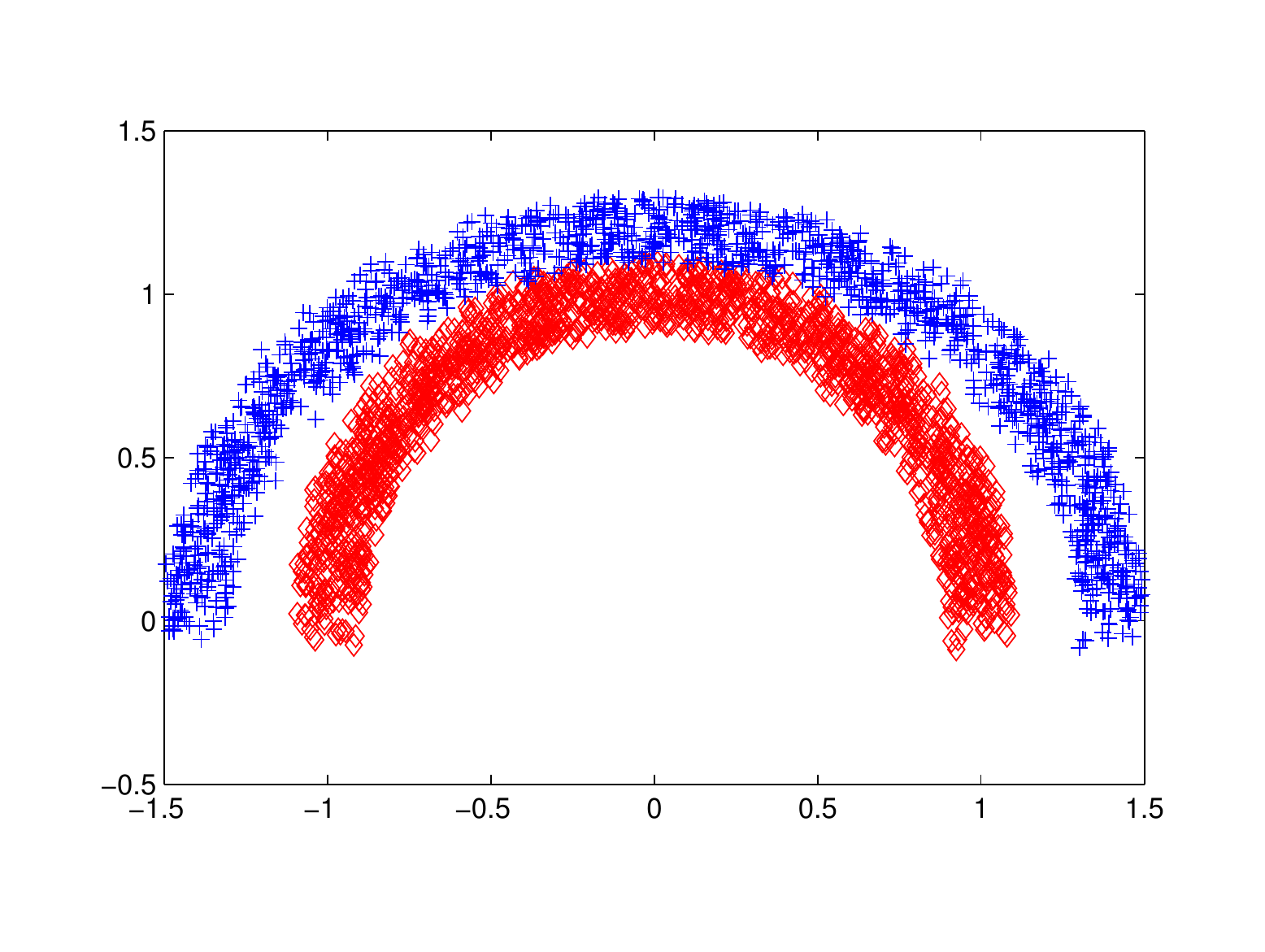}}
  \hfil
\subfigure[Transformed]{\includegraphics[height=1.2in]{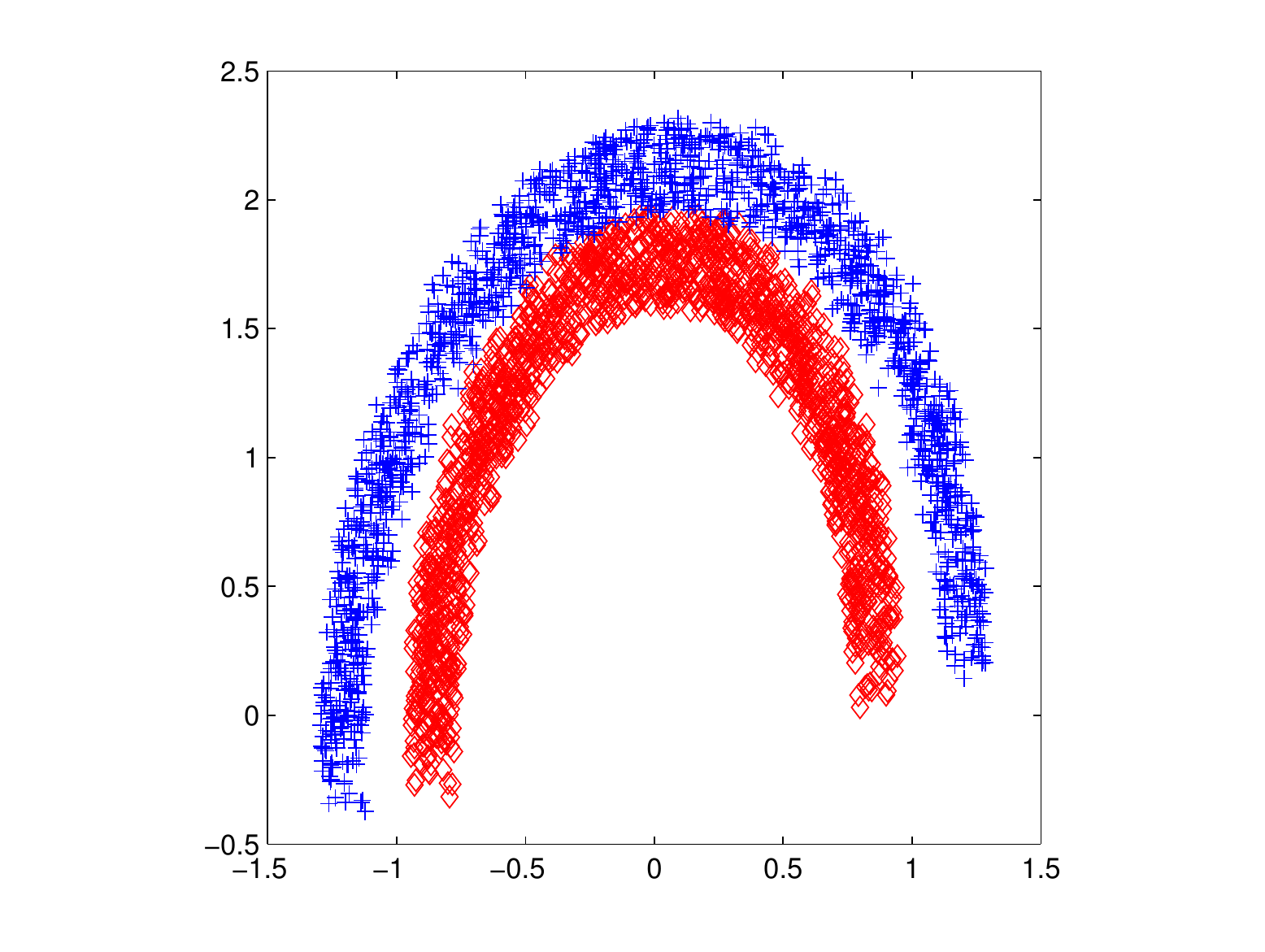}}
\caption{Boundary-based local metric: notice that the critical boundary of two labeled data lies on the top region in left figure, and the right figure depicts the transformed data driven by critical boundary.}
\label{bdlm}
\end{figure}

The idea of learning locally linear distance metrics for kNN classifier is at least 15 years old(DANN\cite{Hastie96}), where linear discriminant analysis is extended to local adaption of the nearest neighbor metric. However, we find few proposals along these lines in literature. How to further justisfy the application of DANN in extension to learn a boundary-based global metrics automatically(maybe in some iterative manner) is still unclear and nontrivial. Figure~\ref{bdlm} depicts the motivation of local linear discriminant analysis, where classical LDA does not work.

In our framework, only the number $k$ of nearest neighbors in kNN is user defined. We only make assumption that the training set reflects the sampling probability density and different sets of class are locally separated, which means in a neighbor, convex hull of points with label of one class has no other points with label of another. Hence the input instance in training set does not necessarily share the same label with its k-nearest neighbors, in other words, the different sets of classes are not necessarily widely separated. Furthermore due to our assumption, the training set should provide information near class boundaries, for the reason that our learning algorithm is supposed to enlarge the margin between different classes.
\subsection{Neighbor estimation}
We estimate the $k$-nearest neighbor distribution of each instance within the same class. For simplicity, we assume it as isotropic Gaussian distribution and obtain the neighbor radius by averaging the distances from its k-nearest neighbors within the same class. We denote the neighbor radius of instance $\mathbf{x}_i$ as $\sigma^{(A)}_i$, which depends on the metric $A$.
\subsection{Offspring model in closed-form convex optimization}
With the estimation of $k$-nearest neighbor within the same class, we expect its neighbors have the same label. Otherwise, we would give a relatively higher penalty weight for neighbor instance with different label by Gaussian filter or Butter-worth filter,
\[p^{(A)}_{i}(\mathbf{x}) = \exp\left(-\dfrac{\norm{\mathbf{x}-\mathbf{x}_i}_A^2}{2(\sigma^{(A)}_i)^2}\right),\]
\[p^{(A)}_{i}(\mathbf{x}) = 1/\left(1+\left({\norm{\mathbf{x}-\mathbf{x}_i}_A}/{\sigma^{(A)}_i}\right)^4\right).\]

We expect to optimize the metric $A$ by diagnosing the non-linear objective
\[\sum_{(i,j)\in K} p^{(A)}_{i}(\mathbf{x}_j)\norm{\mathbf{x}^{(c)}_i-\mathbf{x}_j}_{A}^2\]
with some volume preserving constraints, where template $K$ is a subset of pairs(to be determined later) and $\mathbf{x}^{(c)}_i$ is a ``center'' of ``within class'' neighbor of instance $i$.

Let $A=L^TL$, then above objective can be written as
\[\mbox{Tr}\left(LM_KL^T\right),\]
where
\[M_K = \sum_{(i,j)\in K} p^{(A)}_i(\mathbf{x}_j) (\mathbf{x}^{(c)}_i-\mathbf{x}_j)((\mathbf{x}^{(c)}_i)^T-\mathbf{x}^T_j),\]
here we set weight function $p^{(i)}_{K} = \sum_{(i,j)\in K} p^{(A)}_i(\mathbf{x}_j)$ and normalize $M_K$ by $\overline M_K = M_K/p^{(i)}_{K}$.\\

Considering $D_i=\{(i,j), j\notin C(i)\}$ and $S_i=\{(i,j), j\in C(i)\}$, where $C(i)$ is the set of instances with same label of $\mathbf{x}_i$. Let $N_i = D_i\cup S_i$. We derive another objective sclar function
\[\mathcal{E}(L) = \mbox{Tr}\left(L\left(\sum_{i} w_i \left(\overline M_{D_i}- \overline M_{S_i}\right)\right)L^T\right),\]
where $w_i = p^{(i)}_{D_i}/p^{(i)}_{N_i}$. In our implementation, we select $\mathbf{x}^{(c)}_i = (\sum_{j\in S_i} p^{(A)}_{i}(\mathbf{x}_j)\mathbf{x}_j)/p^{(A)}_{S_i}$ (neighbor-based weighted sum of $S_i$) or $\mathbf{x}^{(c)}_i = \mathbf{x}_i$.\\

The optimization is then to maximize $\mathcal{E}(L)$ with a constraint
\[L\left(\sum_{i} w_i \overline M_{N_i}\right)L^T=I_m\]
where $m$ is the projection dimension. The above optimization problem in fact do have closed form solution by solving a generalized eigenvalue problem (derived from KKT condition):

\[\left(\sum_{i} w_i \left(\overline M_{D_i}- \overline M_{S_i}\right)\right)y_k=\lambda_k\left(\sum_{i} w_i \overline M_{N_i}\right)y_k,\]
where $k=1,\ldots,m\le n$ and $\lambda_1\ge\ldots\ge\lambda_m>0$.

Later we will show the solution is effectively to reduce the amount of $\sum_{i}\log(w_i)$, which could be regarded as a generalization of objective in Neighborhood Component Analysis(NCA\cite{GoldbergerRHS04})

In fact, it is not easy to infer the intrinsic dimension $m$ we should used to project. In empirical experiments, we set $L=[\lambda_1y_1,\ldots,\lambda_my_m]^T$, hence $\lambda_k$ which approximates zero will diminish the contribution of $y_k$.
\subsection{Multiple metric}
Instead of deriving multiple metric geometrically locally for data cluster, we use metric registration approach, which for each point in training set we assign it a link to the final metric sets. For a group of metrics $G={L_0,\ldots,L_s}$, set $A_k=L_k^TL_k$, we derived the offspring process of multiple metrics by modifying

\[w^{(G)}_i = \min_k\{w^{(A_k)}_i\},\]

and the metric associated with single instance $i$ is expected to be the $\mbox{argmin}_k\{w^{(A_k)}_i\}$.

With metric association, the kNN classifier could be extended to multiple metrics intuitively. For a new instance in test set, we count reference instances within the k-nearest neighbor in terms of each metric which have the same metric association, and then select the one which corresponds to the largest count as the test instance's metric association.
\section{Implementation}\label{impl}
There are some key notes in implementation of our overall framework.

In the initial, we pre-compute the a large nearest neighbor $\Omega_i$ for each instance, which in effect assumes $M_{K_i} \approx M_{K_i\cap \Omega_i}$ and $p_{K_i}\approx p_{K_i\cap\Omega_i}$ for $K_i=D_i,S_i,N_i$, and all kNN of $\mathbf{x}_i$ in any metric would fall into $\Omega_i$. Hence our computation complexity is linear during evolution.

During each iteration, we prescribe an active label for each instance when $w_{G}\le \theta$, for some threshold $\theta$, where in experiment we set it equal to $0.1$. Hence inner instances which does not contribute to the critical boundaries will be labeled inactive in the first few evolutionary steps, which in effect improve overall efficiency dramatically.\footnote{Details of CPU time for each method w.r.t dataset should have been provided in terms of chart/table in the final version report.}
\section{Experimental Results}\label{result}
\subsection{Efficiency}
The main computations is kNN classification of validation set, pre-computation of large nearest neighbor $\Omega$ and matrix assembly in closed-form solution maximizing objective $\mathcal{E}$. The former two almost dominate 80\% of the overall computations due to that our framework does not implemented in its most efficient manner in our experiments.

For data set in size smaller than 1000, our implementation works out within seconds, while BoostMetric and LMNN needs 1-2 minutes, and NCA need several minutes. For larger data set(1k-10k), our implementation still only spend no more than 5 minutes, while LMNN and BoostMetric averagely need half an hour or more and NCA is running out of time.

\subsection{UCI dataset}
\begin{table*}[!t]
\renewcommand{\arraystretch}{1.3}
  \caption{UCI dataset error rate(\%) of kNN classification w.r.t different metrics}
  \label{UCIdataset}
{  \centering
  \begin{tabular}{|c|c|c|c|c|c|c|c|c|c|c|}
    \hline
    Data Set  &Euclidean       &PCA         &LDA&NCA&Boost-best&LMNN&CFLML-1&CFLML-3&EM-CFLML&CFLML-best\\\hline
iris
&2.33(2.25)%
&2.67(2.63)%
&\textbf{2.00}(2.81)%
&2.67(3.06)%
&3.33(3.14)%
&\textbf{2.00}(2.81)%
&3.00(2.46)%
&3.67(2.92)%
&3.00(3.31)%
&\textbf{2.00}(1.72)\\\hline
wine
&29.19(7.30)%
&29.46(7.03)%
&\textbf{1.08}(1.40)%
&14.32(8.26)%
&1.62(2.91)%
&5.41(4.23)%
&2.43(2.69)%
&2.43(2.97)%
&2.70(2.85)%
&1.62(2.28)\\\hline
    balance &16.51(1.44)     &14.60(1.51) &8.02(1.85)&5.79(3.67)&8.17(1.63)&13.49(5.46)&6.35(2.21)&5.72(2.21)&5.79(2.08)&\textbf{4.84}(2.03)\\\hline
    wdbc
    &7.74(1.71)
    &7.74(1.71)%
&4.43(0.86)%
&6.61(2.29)%
&\textbf{4.35}(1.83)%
&8.35(2.02)%
&6.09(1.59)%
&5.48(1.69)%
&6.87(2.07)%
&5.30(1.45)\\\hline
vehicle
&32.87(2.66)%
&32.92(2.80)%
&23.04(3.01)%
&25.73(3.23)%
&18.95(2.12)%
&21.40(3.15)%
&19.24(2.00)%
&19.77(3.03)%
&19.88(2.67)%
&\textbf{18.19}(2.03)\\\hline
wine quality
&5.95(0.75)%
&5.95(0.75)%
&\textbf{0.50}(0.23)%
&N/A
&1.07(0.30)%
&1.77(0.47)%
&1.75(1.08)%
&1.73(0.63)%
&1.55(0.57)
&1.33(0.50)\\\hline
    spambase &19.65(1.23) &19.97(0.96)&9.33(0.99)&N/A&18.43(4.37)&10.72(2.70)&7.97(0.75)&8.45(0.62)&7.98(0.59)&\textbf{7.74}(0.54)\\\hline
letters
&4.29
&3.91
&3.86
&N/A
&\textbf{2.82}
&3.14
&3.19
&2.99
&3.19
&2.99\\\hline
isolet
&11.67
&12.76
&4.74
&N/A
&\textbf{4.68}
&5.38
&5.32
&5.32
&4.87
&4.87\\\hline
\end{tabular}
}\fnbelowfloat{isolet are reduced to 100 dimension by PCA}
\end{table*}

We select several data sets from UCI Machine Learning Repository\cite{FrankAsuncion10}, and compare three of our approach(CFLML-1, closed-form metric learning without evolution; CFLML-3, evolution of at most 3 metrics; EM-CFLML, auto-evolution; CFLML-best is the best run of the three in each independent experiment.) in the (multiple metric) kNN classification performance with null(Euclidean distance), Principle Component Analysis(PCA), Linear Discriminant Analysis(LDA\cite{Fisher1962}), Neighborhood Component Analysis(NCA\cite{GoldbergerRHS04}), BoostMetric\cite{ShenKWHJ09}, Large Margin Nearest Neighbor(LMNN\cite{WeinbergerBS05}\cite{WeinbergerS08}).\footnote{The implementation of PCA, LDA and NCA is from Matlab Toolbox for Dimensionality Reduction(\url{http://homepage.tudelft.nl/19j49/Matlab_Toolbox_for_Dimensionality_Reduction.html}), and the code for BoostMetric and LMNN(version 2) is the author's implementation.}

For data set of relevant small size, we run experiment 10 times. Each entry in table~\ref{UCIdataset} represents error means(standard variance) accordingly and N/A denotes running out of time/memory. In each experiment, we randomly divide data set into 80\% for training and 20\% for testing. If validation set is needed, we further cut 15\% out of training set as validation set.

Noting that for LMNN and BoostMetric, we found $k=1,3$ achieves best in their classification performance. But for some data sets, a larger $k$ performs better in null, LDA, and our approach.(For example, we set $k=9$ for wine data set.) In experiments, we select different $k$ individually for each methods to achieve their potentially best performance.

\section{Conclusion and future works}\label{concls}

In this paper, we have introduce a new framework to learn multiple closed-form local metrics(CFLML) for nearest neighbor classification. Given a labeled training set, we have shown how to derived a child metric from a group of parents metrics by solving a closed-form optimization problem. The child metric in some way is supposed and proven to be complementary to parent metrics in their performance of kNN classification. Our framework makes no parametric assumptions about distribution of data and scales naturally, but need to provide a neighbor size $k$ as a trade-off between noise and boundary blurring. By adopting a simple search strategy, multiple metrics and training instances' association are then computed in our framework, and experimental results show that our framework challenges previous single Mahalanobis metric methods in its computational efficiency and classification performance.

In future works, we are interested to refine our closed-form formulation in statistically sound way, optimize the implementation in its efficiency, and design effective evolutionary algorithms.

\appendices




\ifCLASSOPTIONcaptionsoff
  \newpage
\fi



\bibliographystyle{IEEEtran}
%
\bibliography{dml}



%


\begin{IEEEbiography}[{\includegraphics[width=1in,height=1.25in,clip,keepaspectratio]{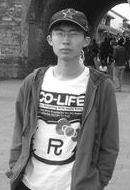}}]{\textsc{Jianbo Ye}}
  received B.S. degree in Mathematics from University of Science and Technology of China, July 2011. His researches are focused on computer graphics, visualization and machine learning.
\end{IEEEbiography}







\end{document}